%% file: 0_main.tex
\documentclass[letterpaper]{article} 
\usepackage{aaai2026}  
\nocopyright
\usepackage{times}  
\usepackage{helvet}  
\usepackage{courier}  
\usepackage[hyphens]{url}  
\usepackage{graphicx} 
\usepackage{amsfonts}
\usepackage{amsmath}
\usepackage{multirow}
\usepackage{booktabs}
\usepackage{xcolor}
\usepackage{enumitem}
\usepackage{natbib}  
\usepackage{caption} 
\usepackage{algorithm}
\usepackage{algorithmic}
\definecolor{lightgray}{rgb}{0.9,0.9,0.9}
\usepackage{pdflscape}
\usepackage[table]{xcolor}
\usepackage{newfloat}
\usepackage{listings}
\DeclareCaptionStyle{ruled}{labelfont=normalfont,labelsep=colon,strut=off} 
\floatstyle{ruled}
\newfloat{listing}{tb}{lst}{}
\floatname{listing}{Listing}
\title{Improving Continual Learning of Knowledge
Graph Embeddings via Informed Initialization}

\author{
    Gerard Pons\textsuperscript{\rm 1},
    Besim Bilalli\textsuperscript{\rm 1},
    Anna Queralt\textsuperscript{\rm 1}
}

\affiliations{
    \textsuperscript{\rm 1}Universitat Politècnica de Catalunya, Barcelona, Catalunya, Spain\\
    gerard.pons.recasens@upc.edu,
    besim.bilalli@upc.edu,
    anna.queralt@upc.edu
}

\usepackage{bibentry}

\begin{document}

\maketitle

\begin{abstract}
Many Knowledege Graphs (KGs) are frequently updated, forcing their Knowledge Graph Embeddings (KGEs) to adapt to these changes. To address this problem, continual learning techniques for KGEs incorporate embeddings for new entities while updating the old ones. One necessary step in these methods is the initialization of the embeddings, as an input to the KGE learning process, which can have an important impact in the accuracy of the final embeddings, as well as in the time required to train them. This is especially relevant for relatively small and frequent updates. We propose a novel informed embedding initialization strategy, which can be seamlessly integrated into existing continual learning methods for KGE, that enhances the acquisition of new knowledge while reducing catastrophic forgetting. Specifically, the KG schema and the previously learned embeddings are utilized to obtain initial representations for the new entities, based on the classes the entities belong to. Our extensive experimental analysis shows that the proposed initialization strategy improves the predictive performance of the resulting KGEs, while also enhancing knowledge retention.  Furthermore, our approach accelerates knowledge acquisition, reducing the number of epochs, and therefore time, required to incrementally learn new embeddings. Finally, its benefits across various types of KGE learning models are demonstrated. 
\end{abstract}

\input{1_introduction}
\input{2_related_work}
\input{3_method}
\input{4_experiments}

\input{5_conclusion}

\bibliography{aaai2026}

\clearpage
\appendix

\input{6_technical}

\end{document}

%% file: 1_introduction.tex
\section{Introduction}
\label{sec:introduction}

Knowledge Graphs (KGs) provide a structured and interconnected way to represent knowledge, by modeling a network of entities and the different relations between them. Information in KGs is represented as a set of triples which describe the interaction of a head and a tail entities via a relation (e.g., [head, relation, tail]). Apart from offering native querying capabilities,  the information contained in the KG can be further utilized in downstream tasks, such as recommender systems \cite{KGREC} or entity classification \cite{KGAPP}. Typically, these Machine Learning (ML) applications expect the input of their models to be vectors. For that, the information stored in the KG can be represented in low-dimensional vectors known as Knowledge Graph Embeddings (KGEs).

KGs grow over time as new triples are added to them. In contexts such as recommender systems, these updates can be relatively small and frequent, and need to be regularly incorporated into KGEs in order to provide timely recommendations. The implications of adding new knowledge to a KGE are twofold. On the one hand, new embeddings must be learned for the entities and relations that were not previously stored in the KG. On the other hand, existing elements in the KG may appear in the new triples, thus their representations should be updated to some extent to capture the new information. In order to incorporate these changes without retraining the model from scratch, continual learning techniques have been proposed. These techniques can be simple and generic such as fine-tuning, or KGE-specific such as LKGE \cite{LKGE} and incDE \cite{incDE}.

\begin{figure}[!t]
\centering
\includegraphics[width=0.4\textwidth]{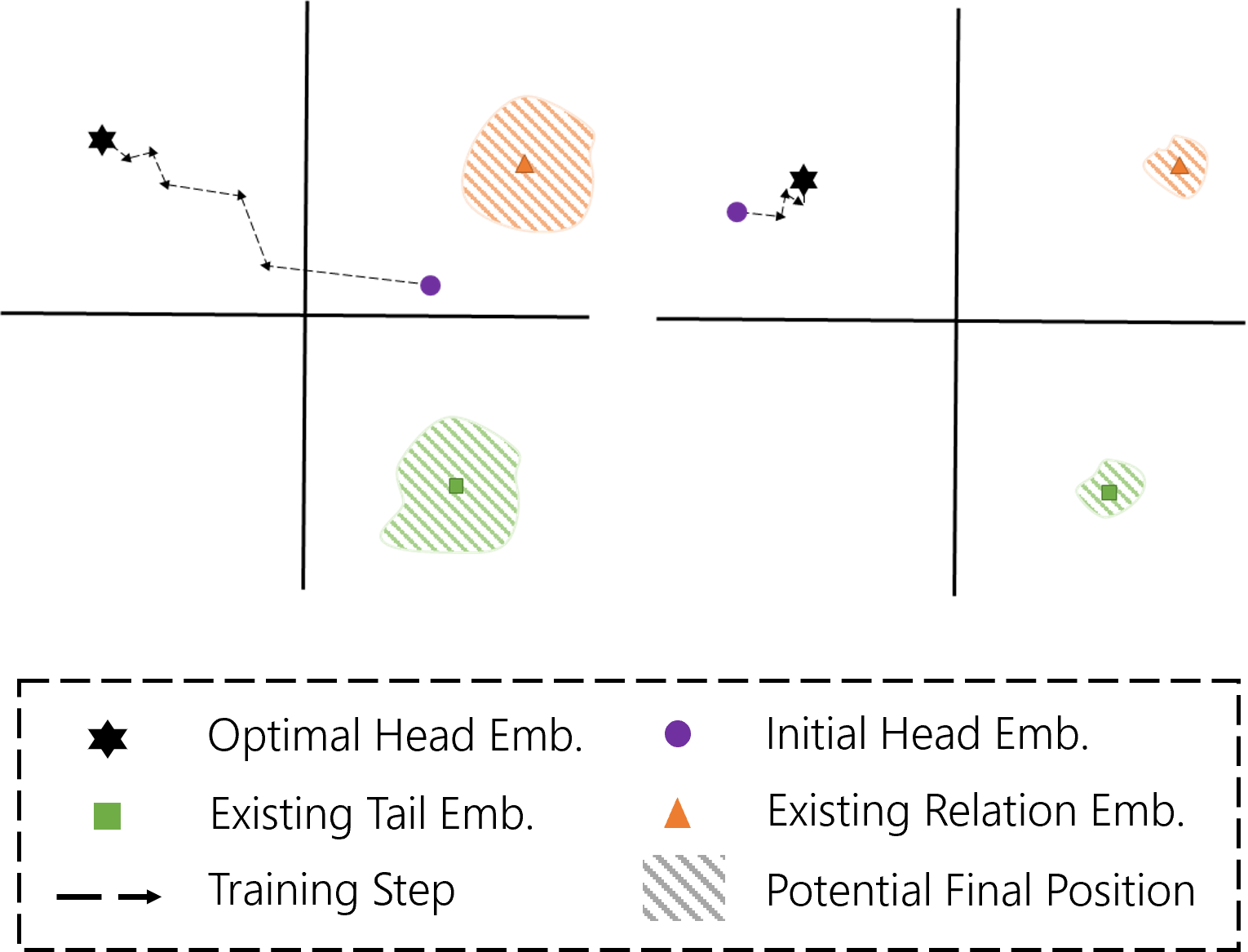}
\caption{Conceptual representation of the potential disruption (represented with a shading area) of existing embeddings when the embedding of a new entity with which they appear in a triple is initialized in a distant (left) or close (right) position from the optimal (left).}
\label{fig1}
\end{figure}

The first step in the incorporation of new information into KGEs is the initialization of the new embeddings. That is, the embeddings of new elements need to have some initial value before starting the learning process. As illustrated in Figure~\ref{fig1}, this initialization has an impact on the training of the embeddings, especially when the initial value of a new embedding is far from its optimal position. First, more steps (thus time) are required for training the new embeddings. Second, if the new triples also include previously existing elements, the training loss for the triple is increased, which translates into adjustments to the embeddings of all the elements in the triple during optimization. Due to these adjustments, existing embeddings may lose information acquired in the past, resulting in catastrophic forgetting \cite{EWC}.

Usually, as an inherited practice from training models from scratch in static settings, the embeddings (or weights in other ML domains) are randomly initialized \cite{incDE, ijcai}, which does not guarantee any proximity to their final position. This has an important impact, not only in the quality of the embedding but also in its required training time. For instance, as will be detailed in the experiments, in some scenarios embeddings can be trained up to nearly 3 times faster and obtain 40\% better predictive performance with an appropriate initialization, rather than relying on random values. In particular, given that continual learning starts from already learned embeddings, one can leverage them for the initialization.

In this work we propose a method for the informed initialization of new embeddings that accelerates the acquisition of new knowledge, while reducing the loss of previously learned information (i.e. catastrophic forgetting). This is particularly relevant in domains where new information arrives frequently and in small increments. Therefore, our focus is on the new entities appearing in this new information (e.g., a new product or user in a recommender system) rather than new relations, as the latter are less frequent and associated with schema changes \cite{kgevolution}.

Our proposal lies in using the schema associated with the KG, together with the already existing KGEs, in order to set the initial values of the new embeddings. In this way, the embeddings inherit the latent information from the classes present in the KGEs, acquiring this knowledge from the beginning. With this initialization, the embeddings can accommodate minor KG updates while preserving their utility, thus postponing the need for a full retraining. Additionally, as initialization is the first step of continual learning techniques (e.g., regularization, distillation or rehearsal methods), our informed initialization can automatically improve the results obtained with these methods in terms of both predictive performance and time.

Therefore, the contributions of this work are as follows:
\begin{itemize}
    \item We propose a method for informed initialization of KGEs in continual learning scenarios, leveraging the KG schema and the existing KGEs. This method enhances both the retention of old knowledge and the acquisition of new one.
    \item We empirically show that the proposed initialization strategy significantly reduces the training time of new embeddings, while obtaining better predictive results.
    \item We introduce metrics to quantify knowledge acquisition and retention in continual learning of KGEs, adapted from incremental task learning in Neural Networks. These metrics are generic and applicable to any continual KGE learning approach, KGE model, and accuracy metric.
    \item We conduct extensive experiments to evaluate the benefits of our informed initialization method across existing continual learning methods and KGE models (e.g. translational and semantic matching models).
    
\end{itemize}

%% file: 2_related_work.tex
\section{Related Work}
\label{sec:related_work}
\subsection{Updating KGEs}
The vast majority of methods that learn KGEs assume a static KG~\cite{KGECO}, thus they are not designed to incrementally consider new triples. 

Inductive learning methods, which leverage known patterns and relationships to generalize and infer new information, have been proposed to obtain the embeddings for new entities. These methods directly compute final embeddings using models such as Graph Neural Networks \cite{MEAN}, attention mechanisms \cite{LAN} or encoders based on structural~\cite{inductiveMetaTransfer, inductiveow} or textual information~\cite{OWE,inductiveDLcompletion}. However, these methods only obtain the new embeddings, and do not update already existing embeddings with the information contained in the new triples. Additionally, they rely on external data, need to learn and maintain additional models, or modify the original KGE model architecture.

These issues can be tackled by using continual learning strategies. Fine-tuning is the most basic one, which involves training over the already learned embeddings with only the new triples, making it much more effective in terms of time and computational costs compared to retraining. However, fine-tuning is prone to catastrophic forgetting \cite{EWC}, as previously learned knowledge is overwritten during training with the new triples.

This problem has been mitigated by extending the fine-tuning strategy with regularization methods, like Elastic Weight Consolidation (EWC), which uses selective penalties to minimize the changes in already learned weights \cite{EWC}, or rehearsal methods, like Episodic Memory Replay (EMR), which samples data with which the model has already been trained and adds it to the new training data \cite{EMR}. Recently, some continual learning methods have been designed specifically for KGEs, namely LKGE \cite{LKGE} or incDE \cite{incDE}. 

As continual learning techniques continue to evolve to address knowledge retention and acquisition challenges, this work presents an initialization strategy that can be integrated with current methods to improve their performance.

\subsection{Initializing Knowledge Graph Embeddings}

In ML models, the weights (e.g., the embeddings in the KGE setting) must be initialized before being adjusted to their final values during the training process. Typically, these weights are randomly initialized from a uniform or Gaussian distribution \cite{goodfellow2016deep}, with Xavier \cite{xavier} or He \cite{he} initialization methods being commonly used to define either the ranges for the uniform distribution or the mean and standard deviation of the Gaussian distribution. These initialization techniques are also used when training KGEs. 

When the KGEs are not trained from scratch but incrementally,  the initialization of the new entities does not need to be random, as it can rely on the already learned embeddings. Some works that focus on the continual learning of KGEs \cite{LKGE, uedkg} have briefly explored this idea. Concretely, they achieve the initialization by using the model’s expected position for the new elements based on the triples they participate in. For instance, in the TransE model \cite{TransE}, in which these approaches are based, the expected position of the embedding of a tail entity is the addition of the relation and head entity embeddings. With this idea, Equation~\ref{equation1} illustrates the initialization proposed in \cite{uedkg}, where $H(e)$ and $T(e)$ are  the triples in which $e$ participates as a head or tail entity, respectively. These triples are composed of a relation and a corresponding tail or head entity, each with an already learned embedding. The functions $f_h$ and $f_t$ obtain the expected position for the entity $e$ according to the model, when it acts as a head or as a tail, respectively. In \cite{LKGE}, the initialization is simplified and only the triples in which $e$ participates as head are considered.
\begin{multline}
\label{equation1}
\mathbf{e} = \frac{1}{|H(e)| + |T(e)|} \Bigg( 
\sum_{(e,r,t) \in H(e)} f_h(r, t) \\
+ \sum_{(h,r,e) \in T(e)} f_t(r, h) 
\Bigg)
\end{multline}

However, these approaches have some drawbacks. First, as pointed out in~\cite{uedkg}, the initialization places the new entities close to the expected position of other elements of the graph, which reduces previous predictive performance. More importantly, these methods rely on the possibility of finding the $f_h$ and $f_t$ functions. This can be trivial for some simple translational embedding methods, such as TransE, but may not be possible for others. For instance, models such as TransH \cite{TransH} or TransF~\cite{TransF} do not use an invertible scoring function which can uniquely identify the expected embedding, while in others such as TransD \cite{TransD}, getting the expected embedding would require solving an underdetermined system of equations. Semantic matching KGE models do not have a geometric relation that should be satisfied, but optimize KGEs by giving higher scores to true triples. Therefore, apart from the requirements of having an invertible scoring function, additional hard constraints must be placed to the norm of the embeddings (e.g.,  $||\mathbf{e}||=1$) to allow finding a solution, which are normally not strictly enforced but encouraged via regularization. The problem is amplified when the KGE model is based on Neural Networks, as apart from having thousands of parameters, making it infeasible to derive a closed-form analytical solution, typically use non-invertible components such as convolutional or activation functions (e.g. ReLU \cite{relu}). This is the case, for instance, of ConvE~\cite{ConvE} and ConvKB \cite{ConvKB}.

Our proposed initialization, however, remains applicable across all models, as it depends only on the generated embeddings and not on the characteristics of the underlying model that generated them.

One could argue that inductive learning approaches could also be used for the initialization of new embeddings. While this is possible, doing so introduces additional complexity, requiring training and maintaining extra models, architecture adjustments, or using external data. In contrast, initialization is regarded as a lightweight step within the training process that integrates with KGE models without requiring architectural modifications or auxiliary training.

%% file: 3_method.tex
\section{Schema-Based Initialization of KGEs}

\subsection{Preliminaries}

A Knowledge Graph (KG) is a collection of triples $\mathcal{T} \subseteq \mathcal{E} \times \mathcal{R} \times \mathcal{E}$, where $\mathcal{E}$ and $\mathcal{R}$ denote the sets of entities and relations, respectively. Since KGs often evolve over time, they can be represented as a sequence of snapshots: $\mathcal{KG} = \{ \mathcal{T}_1, \mathcal{T}_2, \ldots, \mathcal{T}_n\}$, where each snapshot expands on the previous one, i.e., $\mathcal{T}_{i-1} \subseteq \mathcal{T}_i$, $\mathcal{E}_{i-1} \subseteq \mathcal{E}_i$, and $\mathcal{R}_{i-1} \subseteq \mathcal{R}_i$. The newly introduced triples, entities, and relations at snapshot $i$ are denoted as $\Delta \mathcal{T}_i = \mathcal{T}_i \setminus \mathcal{T}_{i-1}$, $\Delta \mathcal{E}_i = \mathcal{E}_i \setminus \mathcal{E}_{i-1}$, and $\Delta \mathcal{R}_i = \mathcal{R}_i \setminus \mathcal{R}_{i-1}$, respectively.
 
KGEs are low-dimensional vector representations of the entities and relations in a KG. A KGE model typically maps each entity $e \in \mathcal{E}$ and relation $r \in \mathcal{R}$ to a vector in a shared $d$-dimensional embedding space. We denote the embedding of an entity $e$ as $\mathbf{e} \in \mathbb{R}^d$ and the embedding of a relation $r$ as $\mathbf{r} \in \mathbb{R}^d$. In the context of non-static KGs, at time step $i$, embeddings must be learned for the newly introduced entities $\Delta \mathcal{E}_i$ and relations $\Delta \mathcal{R}_i$, while the embeddings of existing entities  $\mathcal{E}_{i-1}$ and relations $\mathcal{R}_{i-1}$ need to be updated.

\subsection{Method}

\begin{figure*}[!t]
\centering
\includegraphics[width=\textwidth]{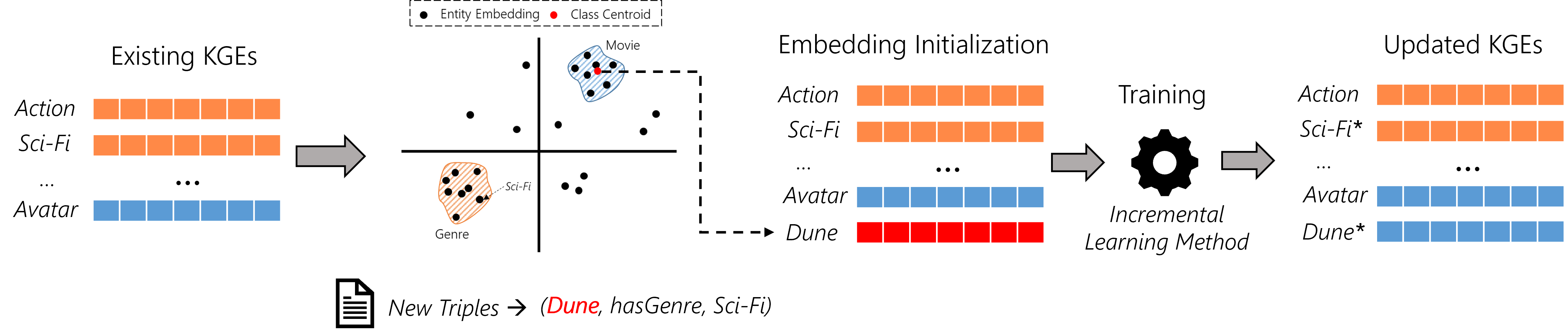}
\caption{Overall process for updating a KGE with new information. Orange and blue vectors represent existing embeddings of entities in the \textit{Genre} and \textit{Movie} classes, respectively. The red vector is the initialized embedding for the new movie \textit{Dune}. The updated KGEs result in a new embedding \textit{Dune*}, and an updated embedding \textit{Sci-Fi*}.}
\label{fig4}
\end{figure*}

We propose a model-agnostic initialization strategy for entity embeddings in non-static KGs, exemplified in Figure~\ref{fig4}. Given a new snapshot at time step $i$ with new triples $\Delta \mathcal{T}_i$, we aim to initialize the embeddings of new entities $\Delta \mathcal{E}_i$ using the KG's schema and previously learned embeddings ${\mathbf{e} \in \mathbb{R}^d : e \in \mathcal{E}_{i-1}}$. These initialized embeddings serve as the input of any continual KGE training method.

The objective of this initialization is to mitigate catastrophic forgetting by positioning new entities closer to their likely final representations, thereby minimizing disruption to existing embeddings $\mathcal{E}_{i-1}$ and reducing training time.

\subsubsection{Schema-based Initialization}
Let $\mathcal{S}$ denote the schema of the KG, which defines the entity types (i.e., classes). Each entity $e \in \mathcal{E}$ is associated with a set of classes $\mathcal{C}_e \subseteq \mathcal{C}$, where $\mathcal{C}$ is the set of all classes defined in $\mathcal{S}$. For each class $c \in \mathcal{C}$, we define its centroid embedding $\mathbf{v}_c$ as:
\begin{equation}
    \mathbf{v}_c = \frac{1}{|\mathcal{E}_c|} \sum_{e \in \mathcal{E}_c} \mathbf{e}, \quad \text{where } \mathcal{E}_c = \{e \in \mathcal{E}_{i-1} : c \in \mathcal{C}_e\}
\end{equation}
Also, we compute the standard deviation $\mathbf{\sigma_c}$ across embeddings of entities in class $c$ to capture intra-class variation.

The embedding of a new entity $e \in \Delta \mathcal{E}_i$ is initialized as the average centroid of the classes it belongs to with a random perturbation:
\begin{equation}
\label{eq:init}
\mathbf{e} = \frac{1}{|\mathcal{C}_e|} \sum_{c \in \mathcal{C}_e} (\mathbf{v}_c + \gamma \cdot \mathbf{\sigma_c} \odot \mathbf{r}_c),
\end{equation}


where $\gamma \in \mathbb{R}^+$ is a hyperparameter controlling the perturbation, $\mathbf{r}_c$ is a random vector sampled independently for each $e$, and $\odot$ denotes element-wise multiplication.

This initialization ensures that entities are initialized with latent information about the classes they belong to, while the stochastic component prevents degenerate initializations (e.g., identical embeddings for entities with identical class memberships), which could lead to undesired behavior~\cite{uedkg}.

Note that the method could be extended to relation embeddings (i.e., by leveraging relation hierarchies from $\mathcal{S}$), although in practical scenarios the introduction of new relations is infrequent and tied to schema changes \cite{kgevolution}.

\subsubsection{Updating KGEs}
Once the embeddings of new entities $e \in \Delta \mathcal{E}_i$ are initialized, any continual learning technique can be used to update the entire embedding space with $\Delta \mathcal{T}_i$.



%% file: 4_experiments.tex
\section{Experiments}
\label{sec:experiments}
This section presents the experimental evaluation of our approach, taking into account different perspectives. We first evaluate the benefits of the proposed initialization method on various continual learning approaches, compared to existing initialization strategies. Then we evaluate the reduction in the required training epochs and time and, finally, we evaluate our initialization strategy on various KGE models. All the implementation details, hyperparameters and datasets are available in the provided Supplementary Material.
\subsection{Experimental Setting}
\subsubsection{Datasets} As mentioned in the Introduction, the focus in on the setting where new information is incrementally added, with each update being relatively small compared to the original KG. To this end, three different datasets have been created from FB15K-237~\cite{fb15k}, a KG completion dataset based on Freebase \cite{freebase}. Following a similar approach to \cite{LKGE, incDE, ijcai}, each dataset consists of an initial set of triples (Snapshot 0), which act as the base KG, and 4 additional sets (Snapshots 1-4) that introduce new entities. Each snapshot is appropriately split into train, validation and test sets. The training set is used to update the embeddings, the validation set for early stopping and the test set to report the results. The three different datasets, named FBinc-\{S,M,L\}, share the same base KG (i.e., Snapshot 0) but have snapshots of different sizes, increasing the number of triples by 2\%, 12\% and 25\% respectively after all the updates. The datasets' characteristics are summarized in Table~\ref{tab:datasets}. The class memberships were obtained by mapping the entities to DBpedia and extracting their associated types from the DBpedia ontology \cite{dbpedia}.

\begin{table}[!t]
\caption{Number of triples in each train snapshot for the different datasets.}
\centering
\begin{tabular}{@{}c|c|c|c@{}}  
\toprule
\textbf{FBinc-} & \textbf{S} & \textbf{M} & \textbf{L}\\ 
\midrule
\textbf{Snapshot 0} & 27832 & 27832 & 27832\\ 
\midrule 
\textbf{Snapshot 1} & 148 & 530 & 1895\\ 
\midrule 
\textbf{Snapshot 2} & 85 & 971 & 1630\\ 
\midrule  
\textbf{Snapshot 3} & 105 & 955 & 1781\\ 
\midrule  
\textbf{Snapshot 4} & 138 & 870 & 1899\\ 
\bottomrule
\end{tabular}
\label{tab:datasets}
\end{table}


\begin{table*}[!t]
\caption{Evaluation of initialization methods  on various continual learning approaches across datasets.}
\centering
\begin{tabular}{@{}ccc|cccc|cccc|cccc@{}}
\toprule
\textbf{Method} & \textbf{Init.} & & \multicolumn{4}{c|}{\textbf{FBinc-S}} & \multicolumn{4}{c|}{\textbf{FBinc-M}} & \multicolumn{4}{c}{\textbf{FBinc-L}} \\
& & & MRR & H@3 & $\Omega_{\text{base}}$ & $\Omega_{\text{new}}$
& MRR & H@3 & $\Omega_{\text{base}}$ & $\Omega_{\text{new}}$
& MRR & H@3 & $\Omega_{\text{base}}$ & $\Omega_{\text{new}}$ \\
\midrule
Retrain & --
& & 0.302 & 0.354 & -- & --
& 0.302 & 0.352 & -- & --
& 0.300 & 0.351 & -- & -- \\
\midrule
\multirow{3}{*}{FT}
& Rand.
& & 0.204 & 0.249 & 0.779 & 0.321
& 0.193 & 0.232 & 0.735 & 0.254
& 0.187 & 0.220 & 0.723 & 0.284 \\
& Model
& & 0.246 & 0.297 & 0.871 & 0.532
& 0.258 & 0.306 & 0.914 & 0.302
& 0.237 & 0.276 & 0.825 & 0.301 \\
& \textbf{Schema}
& & \textbf{0.265} & \textbf{0.316} & \textbf{0.920} & \textbf{0.536}
& \textbf{0.282} & \textbf{0.335} & \textbf{0.964} & \textbf{0.332}
& \textbf{0.268} & \textbf{0.313} & \textbf{0.920} & \textbf{0.329} \\
\midrule
\multirow{3}{*}{EWC}
& Rand.
& & 0.232 & 0.276 & 0.838 & 0.284
& 0.232 & 0.268 & 0.809 & 0.284
& 0.259 & 0.301 & 0.922 & 0.310 \\
& Model
& & 0.264 & 0.312 & 0.912 & 0.465
& 0.274 & 0.323 & 0.948 & 0.273
& 0.265 & 0.310 & 0.905 & 0.358 \\
& \textbf{Schema}
& & \textbf{0.280} & \textbf{0.330} & \textbf{0.953} & \textbf{0.511}
& \textbf{0.286} & \textbf{0.333} & \textbf{0.969} & \textbf{0.314}
& \textbf{0.266} & \textbf{0.311} & \textbf{0.932} & \textbf{0.392} \\
\midrule
\multirow{3}{*}{EMR}
& Rand.
& & 0.202 & 0.244 & 0.771 & 0.311
& 0.196 & 0.232 & 0.746 & 0.239
& 0.186 & 0.217 & 0.715 & 0.276 \\
& Model
& & 0.249 & 0.296 & 0.872 & 0.494
& 0.264 & 0.311 & 0.930 & 0.287
& 0.238 & 0.276 & 0.832 & 0.306 \\
& \textbf{Schema}
& & \textbf{0.267} & \textbf{0.317} & \textbf{0.927} & \textbf{0.514}
& \textbf{0.282} & \textbf{0.330} & \textbf{0.958} & \textbf{0.330}
& \textbf{0.270} & \textbf{0.315} & \textbf{0.928} & \textbf{0.319} \\
\midrule
\multirow{3}{*}{LKGE}
& Rand.
& & 0.243 & 0.285 & 0.844 & 0.279
& 0.257 & 0.294 & 0.830 & 0.317
& 0.286 & 0.333 & 0.953 & 0.340 \\
& Model
& & 0.290 & 0.342 & 0.953 & \textbf{0.552}
& 0.297 & \textbf{0.348} & 0.972 & \textbf{0.363}
& 0.290 & 0.339 & 0.938 & 0.378 \\
& \textbf{Schema}
& & \textbf{0.293} & \textbf{0.343} & \textbf{0.964} & 0.513
& \textbf{0.298} & 0.346 & \textbf{0.978} & 0.331
& \textbf{0.299} & \textbf{0.350} & \textbf{0.962} & \textbf{0.382} \\
\midrule
\multirow{3}{*}{incDE}
& Rand.
& & 0.285 & 0.339 & 0.875 & 0.358
& 0.299 & 0.353 & 0.911 & 0.412
& 0.301 & 0.358 & 0.911 & 0.439 \\
& Model
& & 0.309 & 0.368 & 0.938 & \textbf{0.551}
& 0.311 & 0.369 & 0.925 & 0.463
& \textbf{0.308} & 0.365 & 0.918 & 0.424 \\
& \textbf{Schema}
& & \textbf{0.315} & \textbf{0.376} & \textbf{0.960} & 0.539
& \textbf{0.316} & \textbf{0.378} & \textbf{0.942} & \textbf{0.491}
& \textbf{0.308} & \textbf{0.368} & \textbf{0.926} & \textbf{0.445} \\
\bottomrule
\end{tabular}
\label{tab:exp1}
\end{table*}

\subsubsection{Metrics} The effect of the initialization is evaluated with a link prediction (LP) task, as typically done in KGE evaluation ~\cite{olddog}. LP aims to predict the missing entity in an incomplete triple (e.g., [head, relation, ?] or [?, relation, tail]) by ranking the plausibility of all the entities in the KG for the missing position. The usual LP metrics Mean Reciprocal Rank (MRR) and Hits@3~\cite{KGAPP} are used to evaluate the predictive performance of the resulting embeddings, which will be reported by aggregating the results of all the test sets after training with the final snapshot.

Furthermore, different metrics designed to monitor learning new tasks in Neural Networks \cite{metricscf} have been adapted to monitor knowledge retention and acquisition in KGEs when adding $N$ additional snapshots. These metrics are derived from the previously defined LP performance metrics (Hits@3 in this work), where $\alpha_{i,j}$ represents the value of the evaluation metric on snapshot \textit{i} after training with snapshot \textit{j}: 
\begin{itemize}
    \item \textbf{$\Omega_{base}$} quantifies the retention of old knowledge. It computes the average loss in the evaluation metrics over the base KG every time a new snapshot is added. 
    \begin{equation}
    \label{equation3}
    \Omega_{base} = \frac{1}{N} \sum_{j=1}^S\frac{\alpha_{0,j}}{\alpha_{0,0}}
    \end{equation}
    \item \textbf{$\Omega_{new}$} quantifies the acquisition of new knowledge. It is computed as the average predictive performance of the model after the new snapshots are introduced.
    \begin{equation}
    \label{equation5}
    \Omega_{new} = \frac{1}{N} \sum_{i=1}^S\alpha_{i,i}
    \end{equation}
\end{itemize}

\subsubsection{Initialization Strategies}
Our proposed initialization approach (i.e., \textit{Schema} initialization) has been compared to:
\begin{itemize}
    \item \textit{Random} initialization: This method serves as the baseline and involves initializing the embeddings of new entities by means of Xavier initialization. This is also the mechanism used to initialize entities in Snapshot 0 in all the initialization strategies, as no previous knowledge exists. 
    \item \textit{Model} initialization: Embeddings are initialized with the expected position according to the KGE model \cite{uedkg, LKGE}. Although \textit{Model} initialization is an integral part of these continual approaches, which use TransE as their base KGE model, we evaluate it as a standalone initialization strategy in combination with other continual methods.
\end{itemize}

\subsubsection{Continual Learning Strategies}
To evaluate the generalizability of our proposal, the effect of  initialization has been studied by applying it to different continual learning approaches.

First, and as the most simple method, \textbf{fine-tuning} (FT) has been evaluated. Then, generic continual learning approaches that extend fine-tuning have been examined, namely the regularization method \textbf{EWC} \cite{EWC} and the rehearsal approach \textbf{EMR} \cite{EMR}. Finally, the initialization strategies have been applied to those methods specifically designed for the continual learning of KGEs, concretely \textbf{LKGE} \cite{LKGE} and \textbf{incDE} \cite{incDE}.

\subsection{Effect on Continual Learning Approaches}
\label{sub:exp_init}


Table~\ref{tab:exp1} reports the results of our experiments evaluating the different initialization methods on various continual learning approaches. The base embeddings, shared across continual learning methods and initialization strategies, have been obtained by tuning a TransE model, obtaining the best results with a learning rate of $1e^{-4}$ and an embedding dimension of $200$, as in \cite{LKGE}. Since the \textit{Model} initialization cannot be applied to every KGE model, and because LKGE (the method that proposed the \textit{Model} initialization) and incDE are based on TransE, we use TransE for comparability. For hyperparameter tuning the incremental phase, the learning rate was selected from $\{5e^{-4},1e^{-4},5e^{-5}\}$ and the perturbation parameter $\gamma$ from $\{0,0.1,0.5\}$. Reported results are the average over 5 runs.



The first noteworthy observation is that employing non-random initialization strategies (\textit{Model} and \textit{Schema}) enhances the performance across all metrics, continual learning methods and datasets compared to relying on \textit{Random} initialization. Concretely, a proper initialization is beneficial for the acquisition of new knowledge ($\Omega_{new}$) while at the same time retaining old one ($\Omega_{base}$), being non-regularized methods (i.e., fine-tuning and EMR) the continual learning techniques which benefit the most. For instance, when comparing \textit{Random} and \textit{Schema} initializations, in EMR the retention of base knowledge $\Omega_{base}$ increases by a 20\%, 28\%, and 30\% on the FBinc-S, FBinc-M, and FBinc-L datasets, respectively, while also increases the acquisition of new knowledge $\Omega_{new}$ by 65\%, 38\%, and 16\%  (respectively). Therefore, we observe that the smaller the incremental updates, the more the \textit{Schema} initialization improves the acquisition of new knowledge rather than knowledge retention. Conversely, for larger updates where the new information can more easily be learned, \textit{Schema} initialization has an important role in mitigating catastrophic forgetting compared to the rest of  approaches. Thus, \textit{Schema}  initialization outperforms \textit{Random} and \textit{Model} initializations across dataset sizes when using non-regularized learning methods.

For regularized methods (i.e., EWC, LKGE and incDE), improvements can be seen across all metrics, although the impact becomes less significant with larger incremental updates. Concretely, for LKGE, the continual learning method for which the \textit{Model} initialization was proposed \cite{LKGE}, a similar performance is observed. 

Regarding the $\gamma$ perturbation parameter, it plays a more important role in larger incremental datasets, which can be attributed to reduced embedding degeneration when fewer new entities are added.

\begin{table*}[ht]
\caption{Effect of KGE models and initialization strategies on fine-tuning performance across datasets.}
\centering
\begin{tabular}{@{}ccc|ccc|ccc|ccc@{}}
\toprule
\textbf{Family} & \textbf{Model} & \textbf{Init.} 
& \multicolumn{3}{c|}{\textbf{FBinc-S}} 
& \multicolumn{3}{c|}{\textbf{FBinc-M}} 
& \multicolumn{3}{c}{\textbf{FBinc-L}} \\
& & & H@3 & $\Omega_{\text{base}}$ & $\Omega_{\text{new}}$ 
& H@3 & $\Omega_{\text{base}}$ & $\Omega_{\text{new}}$ 
& H@3 & $\Omega_{\text{base}}$ & $\Omega_{\text{new}}$ \\
\midrule
\multirow{6}{*}{Translational}
& TransH & Random 
& 0.272 & 0.809 & 0.471
& 0.314 & 0.909 & 0.236 
& 0.268 & 0.809 & 0.260 \\
& TransH & Schema 
& \textbf{0.322} & \textbf{0.901} & \textbf{0.516} 
& \textbf{0.343} & \textbf{0.944} & \textbf{0.344} 
& \textbf{0.333} & \textbf{0.918} & \textbf{0.328} \\
\cmidrule{2-12}
& RotatE & Random 
& 0.212 & 0.567 & 0.231
& 0.266 & 0.633 & 0.203 
& 0.300 & 0.714 & 0.359 \\
& RotatE & Schema 
& \textbf{0.351} & \textbf{0.894} & \textbf{0.279} 
& \textbf{0.282} & \textbf{0.796} & \textbf{0.246} 
& \textbf{0.310} & \textbf{0.735} & \textbf{0.431} \\
\cmidrule{2-12}
& TransR & Random 
& 0.152 & 0.719 & \textbf{0.254} 
& 0.139 & 0.645 & \textbf{0.198} 
& 0.086 & 0.438 & 0.232 \\
& TransR & Schema 
& \textbf{0.213} & \textbf{0.901} & 0.245 
& \textbf{0.217} & \textbf{0.936} & 0.193 
& \textbf{0.165} & \textbf{0.717} & \textbf{0.246} \\
\midrule
\multirow{6}{*}{Semantic}
& DistMult & Random 
& 0.268 & 0.782 & 0.257
& 0.317 & 0.921 & 0.146 
& 0.305 & 0.945 & 0.148 \\
& DistMult & Schema 
& \textbf{0.330} & \textbf{0.942} & \textbf{0.274} 
& \textbf{0.346} & \textbf{0.973} & \textbf{0.232} 
& \textbf{0.331} & \textbf{0.975} & \textbf{0.223} \\
\cmidrule{2-12}
& HolE & Random 
& 0.252 & 0.899 & 0.127 
& 0.263 & 0.957 & 0.094 
& 0.235 & 0.938 & 0.110 \\
& HolE & Schema 
& \textbf{0.254} & \textbf{0.915} & \textbf{0.317} 
& \textbf{0.303} & \textbf{0.995} & \textbf{0.280} 
& \textbf{0.277} & \textbf{0.971} & \textbf{0.241} \\
\cmidrule{2-12}
& ProjE & Random 
& 0.351 & \textbf{0.994} & 0.176 
& 0.318 & 0.964 & 0.181 
& 0.271 & \textbf{0.919} & 0.134 \\
& ProjE & Schema 
& \textbf{0.352} & \textbf{0.994} & \textbf{0.238} 
& \textbf{0.326} & \textbf{0.966} & \textbf{0.247} 
& \textbf{0.281} & \textbf{0.919} & \textbf{0.190} \\
\bottomrule
\end{tabular}
\label{tab:kge_init}
\end{table*}

\subsection{Metrics Evolution Over Training Epochs}
\label{sub:exp_epochs}


\begin{figure}[!t]
\centering
\includegraphics[width=0.47\textwidth]{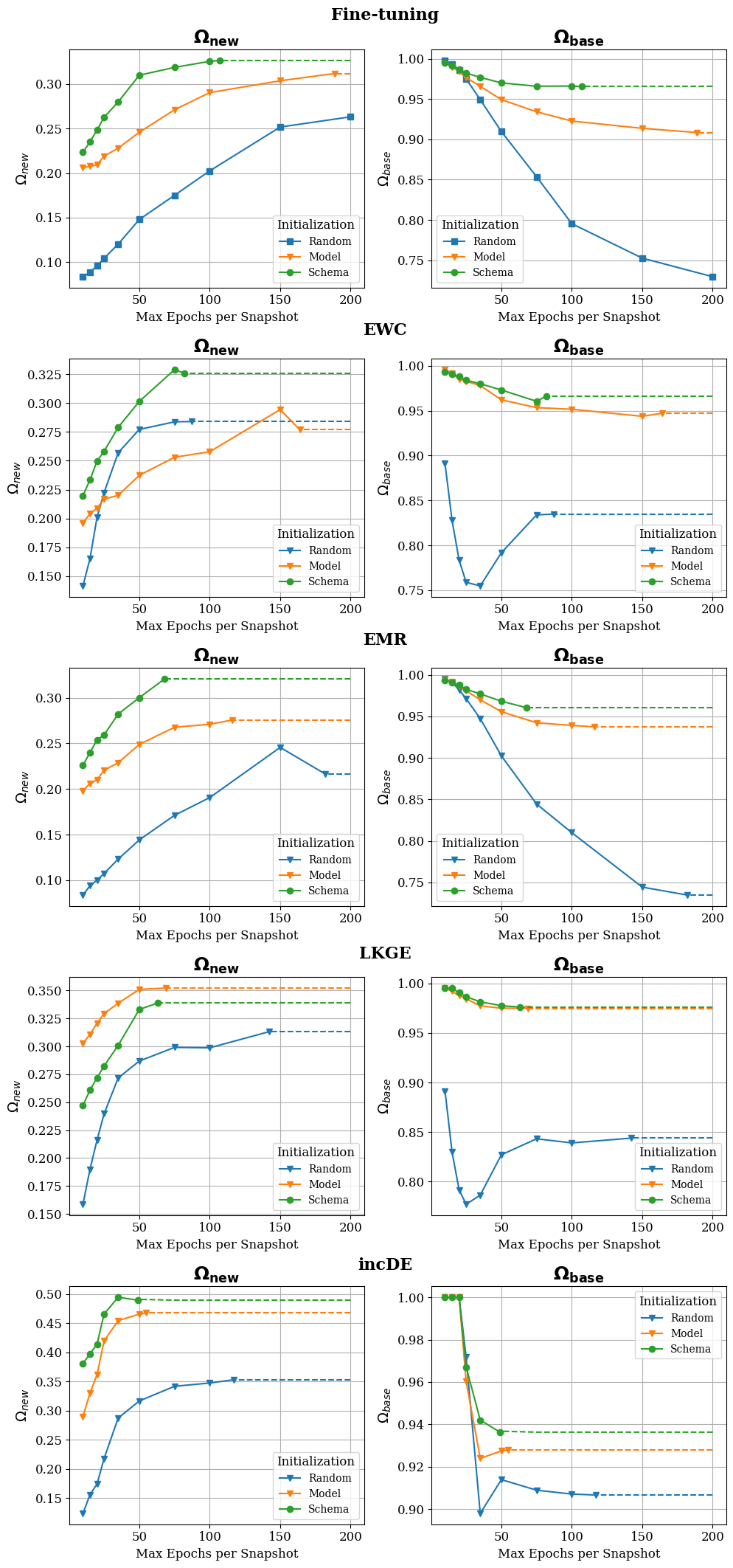}
\caption{Metrics evolution over training epochs on initialization strategies and continual learning methods with the FBinc-M dataset. Dashed lines indicate convergence.}
\label{fig5}
\end{figure}

As exemplified in Figure~\ref{fig1}, a more effective initialization of an embedding can reduce the distance to its final values, thus shortening the training time and minimizing the undesired changes of existing embeddings. To evaluate this hypothesis, an experiment has been conducted in which the number of training epochs for the incremental snapshots has been limited to different values in the range 10 to 200. The results of the experiments are shown in Figure~\ref{fig5} for FBinc-M, the dataset with medium-sized increments, with the best hyperparameters obtained in the previous experiment. Similar results are obtained the FBinc-S and FBinc-L datasets (see the Supplemental Material for details).

For $\Omega_{new}$ with \textit{Schema} initialization, not only more new knowledge is obtained but it is also acquired much faster than in the commonly-used \textit{Random} initialization. In particular, the new knowledge obtained on convergence (see dashed lines) with \textit{Random} initialization is significantly lower (from 6\% in LKGE to 39\% in incDE) than the one obtained with the \textit{Schema} initialization even before convergence (e.g., considering only 50 epochs). Moreover, the \textit{Schema} initialization acquires knowledge with fewer epochs than \textit{Model} initialization, except for the LKGE approach, where the latter is natively integrated. 

As hypothesized,  $\Omega_{base}$ results are linked to the initial effect of the initialization. This is, the more knowledge directly acquired with the initialization, the less knowledge that needs to be obtained from training, thus the less undesired disruption of old embeddings (e.g., see fine-tuning).

Additionally, in all continual approaches the \textit{Schema} initialization converges faster. Compared to \textit{Random} initialization, for all the methods except for EWC where a similar behavior is observed, the number of epochs to convergence is reduced by a factor between 2.16 and 2.67. As this also applies to the other datasets, even when the predictive performance gains are less significant, the reduction in training time makes the \textit{Schema} initialization beneficial, especially as it is applicable to any KGE model.

\subsubsection{Time Analysis}
\label{sub:time}

\begin{figure}
\centering
\includegraphics[width=0.35\textwidth]{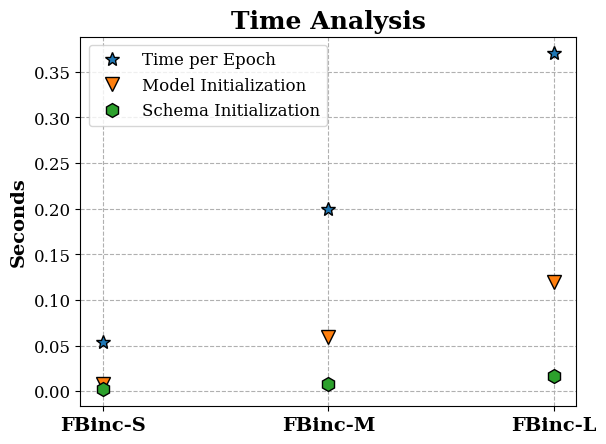}
\caption{Comparison of \textit{Model} initialization and \textit{Schema} initialization with respect to training times for fine-tuning.}
\label{fig9}
\end{figure}

A time analysis has been conducted (see Figure~\ref{fig9}), reporting both the initialization times (we consider the \textit{Random} initialization time to be negligible) and the average time per epoch, using fine-tuning as a lower bound. It can be observed that the \textit{Schema} initialization is faster and has better scalability than the \textit{Model} initialization, and  significantly quicker than a single training epoch. Given that training involves multiple epochs but only a single initialization, and that proper initialization reduces the number of epochs needed to reach convergence, the cost of initialization is not significant in the overall training process. 

Additionally, the size of the base KG only affects in maintaining the centroid metrics (i.e., mean and standard deviation). This operation should be performed between KG updates, thus not impacting the efficiency of the initialization, and requires little computation (e.g., in our case, an order of magnitude less time than a single training epoch). 

\subsection{Different KGE Models}
\label{sub:exp_models}

For comparability purposes, the previous experiments used TransE as the KGE model. This choice was made because the \textit{Model} initialization, originally proposed for TransE, cannot be extended to every model, and because LKGE and incDE are based on TransE. However, given that the proposed \textit{Schema} initialization is model-agnostic, its effectiveness was tested in both translational and semantic KGE models \cite{kge_survey:21}. 


The experiments were conducted by comparing \textit{Schema} and  \textit{Random} initializations using fine-tuning, which allows us to directly leverage models in KGE libraries \cite{pykeen}. The results are shown in Table \ref{tab:kge_init}, which are again obtained by tuning each KGE model (see the Supplementary Material for each specific configuration).

Overall, the models benefit from the \textit{Schema} initialization, which enhances knowledge retention $\Omega_{base}$, acquisition $\Omega_{new}$ or both. Importantly, when only of the metrics improves, the other one is preserved. For instance, in HolE $\Omega_{new}$ more than doubles across all datasets, while $\Omega_{base}$ also improves. In models which experience large amounts of catastrophic forgetting under \textit{Random} initialization (i.e., RotatE and TransR), the \textit{Schema} initialization mitigates these effects. For models that struggle to acquire new knowledge (i.e., ProjE), the \textit{Schema} initialization addresses this limitation while preserving $\Omega_{base}$. Finally, for the best performing model, TransH, $\Omega_{base}$ is improved by a 11\%, 4\% and 13\% and $\Omega_{new}$ by 9\%, 46\% and 26\% for the datasets FBinc-S, FBinc-M and FBinc-L, respectively.

%% file: 5_conclusion.tex
\section{Conclusions}
\label{sec:conclusions}

We introduce a novel and effective entity embedding initialization strategy for KGE continual learning. Our method aims at the quick integration of new information while minimizing disruptions to existing embeddings, by providing a semantically informed starting point for the embeddings of new entities, based on the KG schema. The experiments show that our initialization strategy: i) improves the predictive performance of the downstream task, ii) accelerates the acquisition of new knowledge in continual learning methods, and iii) enhances performance across continual learning strategies and embedding models, making it an effective choice regardless of the specific KGE technique.

%% file: 6_technical.tex
\section*{Appendix A: Initialization Effect on Continual Learning Approaches}

\renewcommand{\thefigure}{A\arabic{figure}}
\renewcommand{\thetable}{A\arabic{table}}

\setcounter{figure}{0}
\setcounter{table}{0}

\begin{table*}[!htp]
\caption{Experiment 1: Final hyperparameter configurations.}
\centering
\begin{tabular}{@{}ccc|cc|cc|cc@{}}
\toprule
\textbf{Method} & \textbf{Init.} & & \multicolumn{2}{c|}{\textbf{FBinc-S}} & \multicolumn{2}{c|}{\textbf{FBinc-M}} & \multicolumn{2}{c}{\textbf{FBinc-L}} \\
& & & Learning Rate & $\gamma$ & Learning Rate & $\gamma$ & Learning Rate & $\gamma$ \\
\midrule
\multirow{3}{*}{FT}
& Random & & $5 \times 10^{-4}$ & -- & $1 \times 10^{-4}$ & -- & $5 \times 10^{-5}$ & -- \\
& Model & & $5 \times 10^{-4}$ & -- & $5 \times 10^{-5}$ & -- & $5 \times 10^{-5}$ & -- \\
& Schema & & $5 \times 10^{-4}$ & 0 & $5 \times 10^{-5}$ & 0 & $5 \times 10^{-5}$ & 0 \\
\midrule
\multirow{3}{*}{EWC}
& Random & & $5 \times 10^{-4}$ & -- &$5 \times 10^{-4}$ & -- & $5 \times 10^{-4}$ & -- \\
& Model & & $5 \times 10^{-4}$ & -- & $5 \times 10^{-5}$ & -- & $5 \times 10^{-4}$ & -- \\
& Schema & & $5 \times 10^{-4}$ & 0 & $5 \times 10^{-5}$ & 0 & $5 \times 10^{-4}$ & 0.1 \\
\midrule
\multirow{3}{*}{EMR}
& Random & & $5 \times 10^{-4}$ & -- & $1 \times 10^{-4}$ & -- & $5 \times 10^{-5}$ & -- \\
& Model & & $5 \times 10^{-4}$ & -- & $5 \times 10^{-5}$ & -- & $5 \times 10^{-5}$ & -- \\
& Schema & & $5 \times 10^{-4}$ & 0 & $5 \times 10^{-5}$ & 0 & $5 \times 10^{-5}$ & 0 \\
\midrule
\multirow{3}{*}{LKGE}
& Random & & $5 \times 10^{-4}$ & -- & $5 \times 10^{-4}$ & -- & $5 \times 10^{-4}$ & -- \\
& Model & & $5 \times 10^{-4}$ & -- & $5 \times 10^{-5}$ & -- & $5 \times 10^{-4}$ & -- \\
& Schema & & $5 \times 10^{-4}$ & 0 & $5 \times 10^{-5}$ & 0 & $5 \times 10^{-4}$ & 0.5 \\
\midrule
\multirow{3}{*}{incDE}
& Random & & $5 \times 10^{-4}$ & -- & $5 \times 10^{-4}$ & -- & $5 \times 10^{-4}$ & -- \\
& Model & & $5 \times 10^{-4}$ & -- & $5 \times 10^{-4}$ & -- & $5 \times 10^{-4}$ & -- \\
& Schema & & $5 \times 10^{-4}$ & 0 & $5 \times 10^{-4}$ & 0 & $5 \times 10^{-4}$ & 0.1 \\
\bottomrule
\end{tabular}
\label{tab:exp1}
\end{table*}

\begin{table*}[!t]
\caption{Effect of different initialization strategies on FBinc-S (mean $\pm$ std). In gray, the metrics for which \textit{Schema} initialization has an average greater performance but cannot be said to be significant are highlighted.}
\centering
\begin{tabular}{@{}ccc|cccc@{}}
\toprule
\textbf{Method} & \textbf{Init.} & & MRR & H@3 & $\Omega_{\text{base}}$ & $\Omega_{\text{new}}$ \\
\midrule
\multirow{3}{*}{FT}
& Rand.   & & 0.204 $\pm$ 0.003 & 0.249 $\pm$ 0.004 & 0.779 $\pm$ 0.009 & 0.321 $\pm$ 0.019 \\
& Model   & & 0.246 $\pm$ 0.000 & 0.297 $\pm$ 0.001 & 0.871 $\pm$ 0.007 & \cellcolor{lightgray} 0.532 $\pm$ 0.13 \\
& Schema  & & \textbf{0.265 $\pm$ 0.003} & \textbf{0.316 $\pm$ 0.004} & \textbf{0.920 $\pm$ 0.012} & \textbf{0.536 $\pm$ 0.009} \\
\midrule
\multirow{3}{*}{EWC}
& Rand.   & & 0.232 $\pm$ 0.002 & 0.276 $\pm$ 0.003 & 0.838 $\pm$ 0.003 & 0.284 $\pm$ 0.011 \\
& Model   & & 0.264 $\pm$ 0.001 & 0.312 $\pm$ 0.002 & 0.912 $\pm$ 0.004 & 0.465 $\pm$ 0.010 \\
& Schema  & & \textbf{0.280 $\pm$ 0.001} & \textbf{0.330 $\pm$ 0.001} & \textbf{0.953 $\pm$ 0.003} & \textbf{0.511 $\pm$ 0.020} \\
\midrule
\multirow{3}{*}{EMR}
& Rand.   & & 0.202 $\pm$ 0.002 & 0.244 $\pm$ 0.002 & 0.771  $\pm$ 0.005& 0.311 $\pm$ 0.013\\
& Model   & & 0.249 $\pm$ 0.003 & 0.296 $\pm$ 0.003 & 0.872 $\pm$ 0.006 & \cellcolor{lightgray} 0.494 $\pm$ 0.023 \\
& Schema  & & \textbf{0.267 $\pm$ 0.006} & \textbf{0.317 $\pm$ 0.006} & \textbf{0.927 $\pm$ 0.011} & \textbf{0.514 $\pm$ 0.023} \\
\midrule
\multirow{3}{*}{LKGE}
& Rand.   & & 0.243 $\pm$ 0.004 & 0.285 $\pm$ 0.003 & 0.844 $\pm$ 0.005 & 0.279 $\pm$ 0.003 \\
& Model   & & \cellcolor{lightgray} 0.290 $\pm$ 0.002 & \cellcolor{lightgray} 0.342 $\pm$ 0.003 & 0.953 $\pm$ 0.006 & \textbf{0.552 $\pm$ 0.015} \\
& Schema  & & \textbf{0.293 $\pm$ 0.002} & \textbf{0.343 $\pm$ 0.001} & \textbf{0.964 $\pm$ 0.004} & 0.513 $\pm$ 0.014 \\
\midrule
\multirow{3}{*}{incDE}
& Rand.   & & 0.285 $\pm$ 0.002 & 0.339 $\pm$ 0.004 & 0.875 $\pm$ 0.003& 0.358 $\pm$ 0.024 \\
& Model   & & 0.309 $\pm$ 0.002 & 0.368 $\pm$ 0.003& 0.938 $\pm$ 0.003 & \textbf{0.551 $\pm$ 0.006} \\
& Schema  & & \textbf{0.315 $\pm$ 0.002} & \textbf{0.376 $\pm$ 0.002} & \textbf{0.960 $\pm$ 0.005} & 0.539 $\pm$ 0.006 \\
\bottomrule
\end{tabular}
\label{tab:fbincs}
\end{table*}

\begin{table*}[!t]
\caption{Effect of different initialization strategies on FBinc-M (mean $\pm$ std). In gray, the metrics for which \textit{Schema} initialization has an average greater performance but cannot be said to be significant are highlighted.}
\centering
\begin{tabular}{@{}ccc|cccc@{}}
\toprule
\textbf{Method} & \textbf{Init.} & & MRR & H@3 & $\Omega_{\text{base}}$ & $\Omega_{\text{new}}$ \\
\midrule
\multirow{3}{*}{FT}
& Rand.   & & 0.193 $\pm$ 0.005 & 0.232 $\pm$ 0.005 & 0.735 $\pm$ 0.008 & 0.254 $\pm$ 0.014 \\
& Model   & & 0.258 $\pm$ 0.002 & 0.306 $\pm$ 0.002 & 0.914 $\pm$ 0.010 & 0.302 $\pm$ 0.011 \\
& Schema  & & \textbf{0.282 $\pm$ 0.001} & \textbf{0.335 $\pm$ 0.001} & \textbf{0.964 $\pm$ 0.007} & \textbf{0.332 $\pm$ 0.017} \\
\midrule
\multirow{3}{*}{EWC}
& Rand.   & & 0.232 $\pm$ 0.004 & 0.268 $\pm$ 0.005 & 0.809 $\pm$ 0.015& 0.284 $\pm$ 0.007 \\
& Model   & & 0.274 $\pm$ 0.002 & 0.323 $\pm$ 0.002 & 0.948 $\pm$ 0.006& 0.273 $\pm$ 0.010 \\
& Schema  & & \textbf{0.286 $\pm$ 0.001} & \textbf{0.333 $\pm$ 0.001} & \textbf{0.969 $\pm$ 0.008} & \textbf{0.314 $\pm$ 0.019} \\
\midrule
\multirow{3}{*}{EMR}
& Rand.   & & 0.196 $\pm$ 0.004 & 0.232 $\pm$ 0.004 & 0.746 $\pm$ 0.010 & 0.239 $\pm$ 0.022 \\
& Model   & & 0.264 $\pm$ 0.005 & 0.311 $\pm$ 0.006 & 0.930 $\pm$ 0.014 & 0.287 $\pm$ 0.018 \\
& Schema  & & \textbf{0.282 $\pm$ 0.001} & \textbf{0.330 $\pm$ 0.002} & \textbf{0.958 $\pm$ 0.006} & \textbf{0.330 $\pm$ 0.008} \\
\midrule
\multirow{3}{*}{LKGE}
& Rand.   & & 0.257 $\pm$ 0.004 & 0.294 $\pm$ 0.006 & 0.830 $\pm$ 0.0011 & \cellcolor{lightgray} 0.317 $\pm$ 0.008 \\
& Model   & & \cellcolor{lightgray} 0.297 $\pm$ 0.001 & \textbf{0.348 $\pm$ 0.001} & \cellcolor{lightgray} 0.972 $\pm$ 0.003 & \textbf{0.363 $\pm$ 0.012} \\
& Schema  & & \textbf{0.298 $\pm$ 0.001} & 0.346 $\pm$ 0.001 & \textbf{0.978 $\pm$ 0.005} & 0.331 $\pm$ 0.017 \\
\midrule
\multirow{3}{*}{incDE}
& Rand.   & & 0.299 $\pm$ 0.003 & 0.353 $\pm$ 0.004 & 0.911 $\pm$ 0.006 & 0.412 $\pm$ 0.021 \\
& Model   & & 0.311 $\pm$ 0.001 & 0.369 $\pm$ 0.002 & 0.925 $\pm$ 0.003 & 0.463 $\pm$ 0.004 \\
& Schema  & & \textbf{0.316 $\pm$ 0.001} & \textbf{0.378 $\pm$ 0.002} & \textbf{0.942 $\pm$ 0.004} & \textbf{0.491 $\pm$ 0.006} \\
\bottomrule
\end{tabular}
\label{tab:fbincm}
\end{table*}

\begin{table*}[!t]
\caption{Effect of different initialization strategies on FBinc-L (mean $\pm$ std). In gray, the metrics for which \textit{Schema} initialization has an average greater performance but cannot be said to be significant are highlighted.}
\centering
\begin{tabular}{@{}ccc|cccc@{}}
\toprule
\textbf{Method} & \textbf{Init.} & & MRR & H@3 & $\Omega_{\text{base}}$ & $\Omega_{\text{new}}$ \\
\midrule
\multirow{3}{*}{FT}
& Rand.   & & 0.187 $\pm$ 0.003 & 0.220 $\pm$ 0.004 & 0.723 $\pm$ 0.006 & 0.284 $\pm$ 0.005\\
& Model   & & 0.237 $\pm$ 0.002 & 0.276 $\pm$ 0.003 & 0.825 $\pm$ 0.013 & 0.301 $\pm$ 0.005 \\
& Schema  & & \textbf{0.268 $\pm$ 0.001} & \textbf{0.313 $\pm$ 0.002} & \textbf{0.920 $\pm$ 0.009} & \textbf{0.329 $\pm$ 0.003} \\
\midrule
\multirow{3}{*}{EWC}
& Rand.   & & \cellcolor{lightgray} 0.259 $\pm$ 0.007 & 0.301 $\pm$ 0.010 & 0.922 $\pm$ 0.020 & 0.310 $\pm$ 0.008 \\
& Model   & & \cellcolor{lightgray} 0.265 $\pm$ 0.002 & \cellcolor{lightgray} 0.310 $\pm$ 0.003 & 0.905 $\pm$ 0.006 & 0.358 $\pm$ 0.008 \\
& Schema  & & \textbf{0.266 $\pm$ 0.002} & \textbf{0.311 $\pm$ 0.001} & \textbf{0.932 $\pm$ 0.004} & \textbf{0.392 $\pm$ 0.002} \\
\midrule
\multirow{3}{*}{EMR}
& Rand.   & & 0.186 $\pm$ 0.002 & 0.217 $\pm$ 0.003 & 0.715 $\pm$ 0.006 & 0.276 $\pm$ 0.002 \\
& Model   & & 0.238 $\pm$ 0.002 & 0.276 $\pm$ 0.003 & 0.832 $\pm$ 0.009 & 0.306 $\pm$ 0.004 \\
& Schema  & & \textbf{0.270 $\pm$ 0.003} & \textbf{0.315 $\pm$ 0.004} & \textbf{0.928 $\pm$ 0.011} & \textbf{0.319 $\pm$ 0.008} \\
\midrule
\multirow{3}{*}{LKGE}
& Rand.   & & 0.286 $\pm$ 0.004 & 0.333 $\pm$ 0.006 & 0.953 $\pm$ 0.012 & 0.340 $\pm$ 0.013 \\
& Model   & & 0.290 $\pm$ 0.002 & 0.339 $\pm$ 0.003 & 0.938 $\pm$ 0.006 & 0.378 $\pm$ 0.003 \\
& Schema  & & \textbf{0.299 $\pm$ 0.002} & \textbf{0.350 $\pm$ 0.003} & \textbf{0.962 $\pm$ 0.007} & \textbf{0.382 $\pm$ 0.005} \\
\midrule
\multirow{3}{*}{incDE}
& Rand.   & & 0.301 $\pm$ 0.001 & 0.358 $\pm$ 0.002 & \cellcolor{lightgray} 0.911 $\pm$ 0.003 & \cellcolor{lightgray} 0.439 $\pm$ 0.005 \\
& Model   & & \textbf{0.308 $\pm$ 0.001} & \cellcolor{lightgray} 0.365 $\pm$ 0.001 & \cellcolor{lightgray} 0.918 $\pm$ 0.002 & 0.424 $\pm$ 0.005 \\
& Schema  & & \textbf{0.308 $\pm$ 0.001} & \textbf{0.368 $\pm$ 0.003} & \textbf{0.926 $\pm$ 0.007} & \textbf{0.445 $\pm$ 0.005} \\
\bottomrule
\end{tabular}
\label{tab:fbincl}
\end{table*}

All the experiments start from the same embeddings of the base snapshot, learned from TransE. As done in \cite{LKGE}, the base model hyperparameters have been tuned with a learning rate in $\{0.001,0.0005,0.0001\}$, a batch size in $\{1024, 2048\}$ and an embedding dimension in $\{100, 200\}$. All the experiments have been run in a GeForce RTX 4080 SUPER with 64GB RAM and 2 TB SSD. The best model has been obtained with the following hyperparameters:
\begin{itemize}
    \item \textbf{Embedding Dimension}: 200
    \item \textbf{Batch Size}: 2048
    \item \textbf{Learning Rate}: 0.0001
\end{itemize}

The results obtained for this model are an MRR of 0.309 and Hits@3 of 0.358.

Additionally, some parameters have been fixed:
\begin{itemize}
    \item \textbf{Number of Negative Samples}: The number of corrupt triples against which every positive triple is compared to has been set to 10.
    \item \textbf{Patience}: The embeddings have been trained with early stopping, with a patience of 3 epochs on the MRR over the validation set.
\end{itemize}

For the continual learning setting, the following hyperparameters have been tested in a grid-search:
\begin{itemize}
    \item \textbf{Learning Rate}: The learning rate has been set in $\{0.0005,0.0001,0.00005\}$, which include the optimal rate for training the base KGE, as well as higher and lower rates.
    \item \textbf{Random Seed}: The results are reported as an average of 5 runs, having the random seed set in $\{11,22,33,44,55\}$.
    \item \textbf{Perturbation Parameter}: The noise added to the initialization is controlled with the parameter $\gamma$, which is set in $\{0,0.1,0.5\}$. This allows having updates which are not disturbed by the randomness.
\end{itemize}

The final hyperparameter configurations can be seen in Table \ref{tab:exp1}.

Additionally, in Table \ref{tab:fbincs}, Table \ref{tab:fbincm} and Table \ref{tab:fbincl}  the experiment results are shown with the mean and standard deviations for the FBinc-S, FBinc-M and FBinc-L datasets, respectively. Also, a Wilcoxon test has been perform to assess if, for the metrics where \textit{Schema} initialization has better results, those are significantly better than the ones from \textit{Random} and \textit{Model} initializations. In light gray, the metrics for which that does not apply (i.e., $p{\text-}value \geq 0.05$) are highlighted.  

\clearpage
\section*{Appendix B: Metrics Evolution Over Training Epochs}

\renewcommand{\thefigure}{B\arabic{figure}}
\renewcommand{\thetable}{B\arabic{table}}

\setcounter{figure}{0}
\setcounter{table}{0}

\begin{figure}[!ht]
\centering
\includegraphics[width=0.46\textwidth]{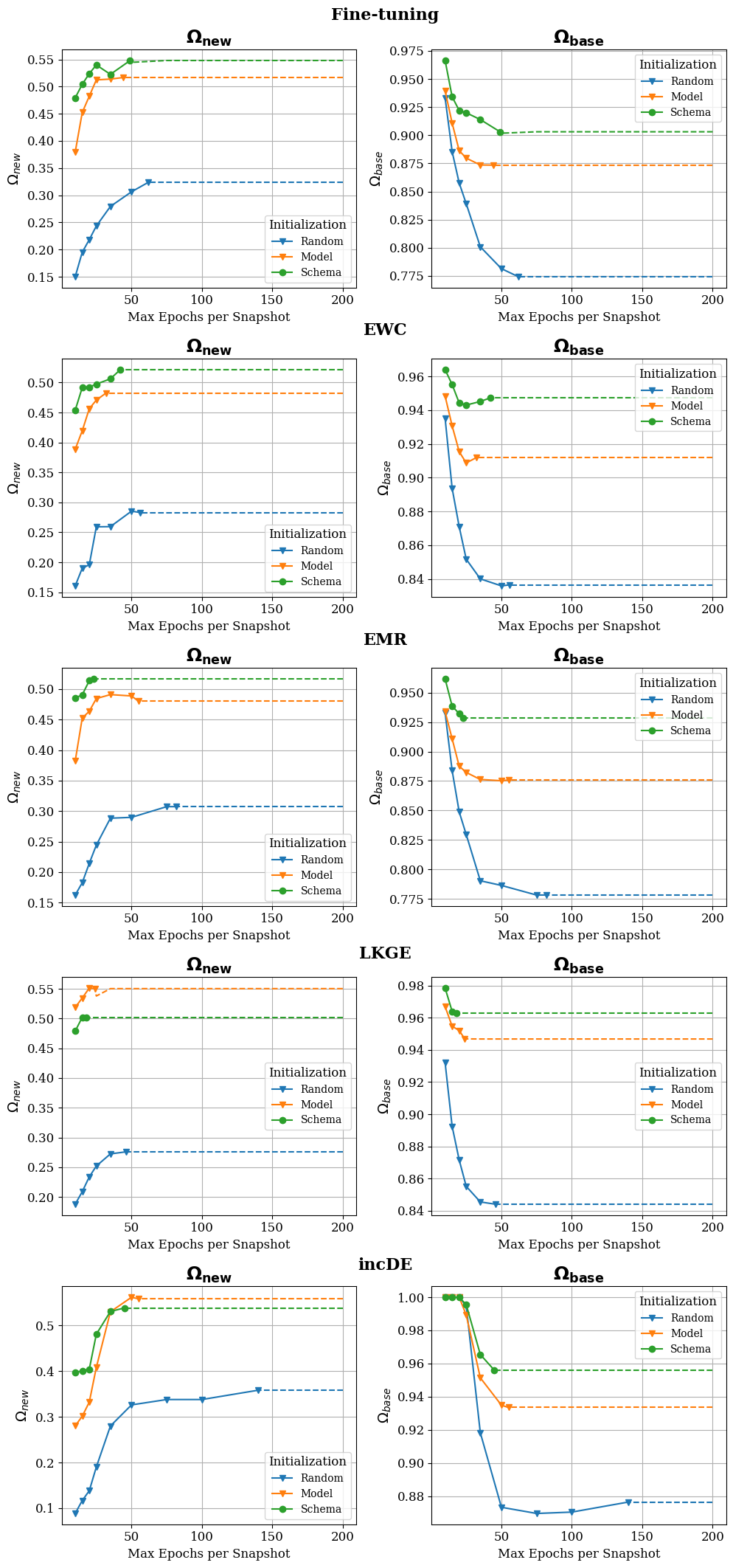}
\caption{Impact of training epochs on various initialization strategies with the FBinc-S dataset. Dashed lines indicate convergence.}
\label{fig1}
\end{figure}

\begin{figure}[!ht]
\centering
\includegraphics[width=0.48\textwidth]{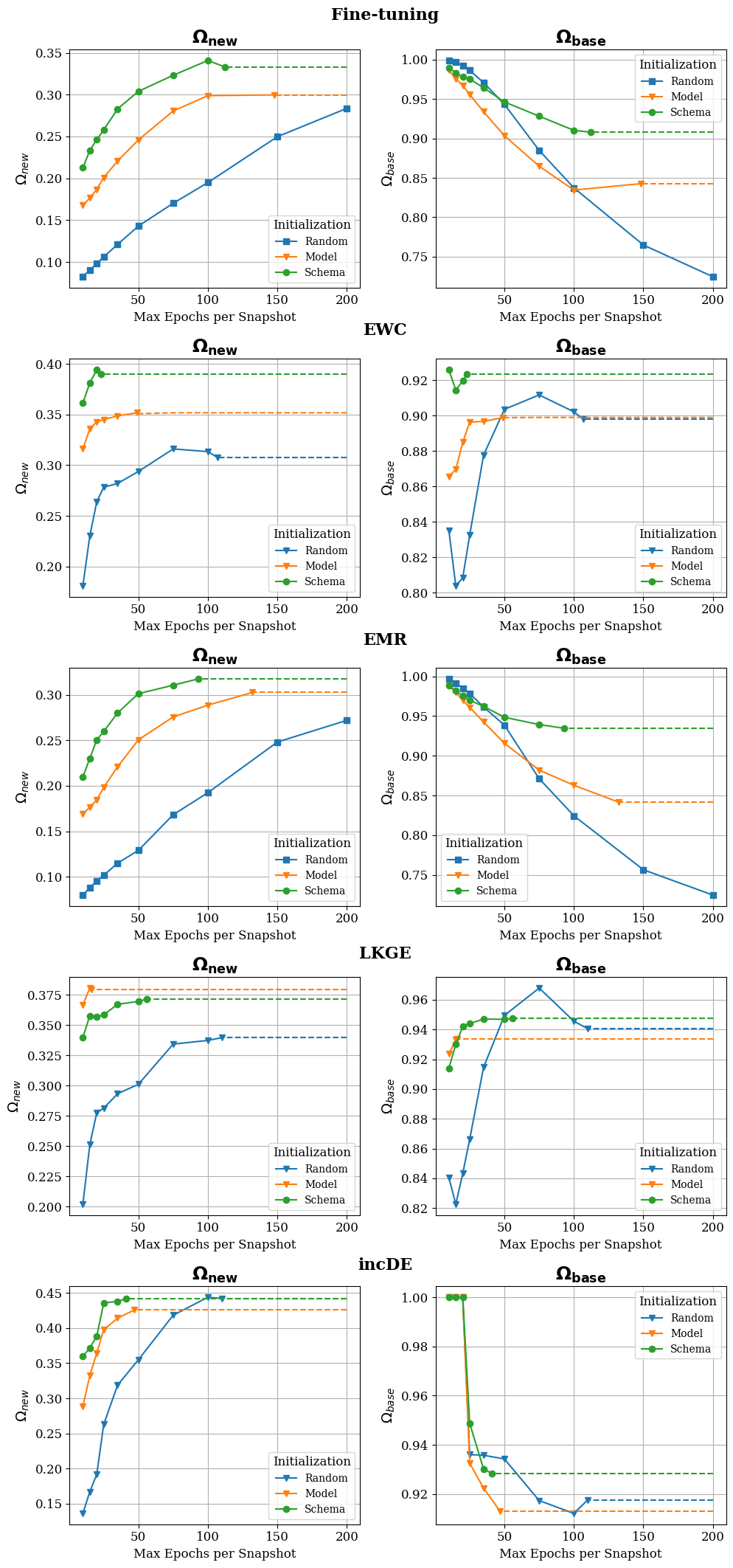}
\caption{Impact of training epochs on various initialization strategies with the FBinc-L dataset. Dashed lines indicate convergence.}
\label{fig2}
\end{figure}

In the second experiment, the impact of the number of allowed epoch was studied, by using the configuration that obtained the best hyperparameters for each dataset and continual learning method in Experiment 1. In Figure \ref{fig1} and Figure \ref{fig2}, the results for the FBinc-S and FBinc-L datasets are shown, respectively.

\clearpage
\section*{Appendix C: Different KGE models}

\renewcommand{\thefigure}{C\arabic{figure}}
\renewcommand{\thetable}{C\arabic{table}}

\setcounter{figure}{0}
\setcounter{table}{0}

\begin{table*}[!ht]
\caption{Experiment 3: Final hyperparameter configurations for the continual learning technique learning rate and the $\gamma$ perturbation parameter.}
\centering
\begin{tabular}{@{}ccc|cc|cc|cc@{}}
\toprule
\textbf{Family} & \textbf{Model} & \textbf{Init.} 
& \multicolumn{2}{c|}{\textbf{FBinc-S}} 
& \multicolumn{2}{c|}{\textbf{FBinc-M}} 
& \multicolumn{2}{c}{\textbf{FBinc-L}} \\
& & & Learning Rate & $\gamma$ 
&  Learning Rate & $\gamma$ 
&  Learning Rate & $\gamma$ \\
\midrule
\multirow{6}{*}{Translational}
& TransH & Random 
& $5 \times 10^{-4}$ & --
& $1 \times 10^{-4}$ & -- 
& $1 \times 10^{-4}$ & -- \\
& TransH & Schema 
& $1 \times 10^{-3}$ & 0 
& $1 \times 10^{-4}$ & 0 
& $1 \times 10^{-4}$ & 0 \\
\cmidrule{2-9}
& RotatE & Random 
& $1 \times 10^{-3}$ & --
& $5 \times 10^{-4}$ & -- 
& $1 \times 10^{-3}$ & -- \\
& RotatE & Schema 
& $1 \times 10^{-3}$ & 1 
& $1 \times 10^{-3}$ & 1 
& $1 \times 10^{-3}$ & 1 \\
\cmidrule{2-9}
& TransR & Random 
& $5 \times 10^{-5}$ & -- 
& $5 \times 10^{-6}$ & -- 
& $5 \times 10^{-6}$ & -- \\
& TransR & Schema 
& $5 \times 10^{-5}$ & 0 
& $1 \times 10^{-5}$ & 0 
& $5 \times 10^{-6}$ & 0 \\
\midrule
\multirow{6}{*}{Semantic}
& DistMult & Random 
& $5 \times 10^{-4}$ & --
& $1 \times 10^{-4}$ & -- 
& $1 \times 10^{-4}$ & -- \\
& DistMult & Schema 
& $1 \times 10^{-3}$ & 0 
& $1 \times 10^{-4}$ & 0.1 
& $1 \times 10^{-4}$ & 0.1 \\
\cmidrule{2-9}
& HolE & Random 
& $5 \times 10^{-4}$ & -- 
& $1 \times 10^{-4}$ & -- 
& $1 \times 10^{-4}$ & -- \\
& HolE & Schema 
& $5 \times 10^{-4}$ & 0.1 
& $1 \times 10^{-4}$ & 0.1 
& $1 \times 10^{-4}$ & 0.5 \\
\cmidrule{2-9}
& ProjE & Random 
& $1 \times 10^{-4}$ & -- 
& $1 \times 10^{-4}$ & -- 
& $1 \times 10^{-4}$ & -- \\
& ProjE & Schema 
& $1 \times 10^{-4}$ & 0 
& $1 \times 10^{-4}$ & 0.1 
& $1 \times 10^{-4}$ & 0.5 \\
\bottomrule
\end{tabular}
\label{tab:exp2}
\end{table*}

For testing the effect of the initialization over different KGE models, the Python library PyKEEN \cite{pykeen} has been used, which provides implementations for KGE models. For each model, the same base KGE have been obtained, from which continual learning will start. The base embeddings have been obtained by using the Adam optimizer, a number of negative samples of 10, the default settings for early stopping and SLWA training method ~\cite{olddog}, the embedding dimension of 200 and a learning rate from $\{0.001,0.0005,0.0001\}$. For all models the learning rate for the best base KGE has been of 0.001, except for TransR, which the best learning rate has been obtained with 0.0001.

For the continual learning setting, the following hyperparameters have been tested in a grid-search:
\begin{itemize}
    \item \textbf{Learning Rate}: The learning rate has been set in $\{0.0005, 0.001,0.0005,0.0001\}$. However, for TransR which resulted to be very sensitive to continual learning, the rates have been lowered to $\{0.00005,0.000001,0.000005\}$.
    
    \item \textbf{Random Seed}: The results are reported as an average of 5 runs, having the random seed set in $\{11,22,33,44,55\}$.
    \item \textbf{Perturbation Parameter}: The noise added to the initialization is controlled with the parameter $\gamma$, which is set in $\{0,0.1,0.5,1\}$. 
    
\end{itemize}

The final hyperparameter configurations are reported in Table \ref{tab:exp2}. Additionally, in Table \ref{tab:kges}, Table \ref{tab:kgem} and Table \ref{tab:fbincl}, the experiment results are shown with the mean and standard deviations for the FBinc-S, FBinc-M and FBinc-L datasets, respectively. Finally, a Wilcoxon test has been perform to assess where the \textit{Schema} initialization is significantly better than the \textit{Random} initialization. In light gray, the metrics for which that does not apply (i.e., $p{\text-}value \geq 0.05$) are highlighted.

\clearpage

\begin{table*}[ht]
\caption{Fine-tuning performance on \textbf{FBinc-S} dataset (mean $\pm$ std). In gray, the metrics for which \textit{Schema} initialization has an average greater performance but cannot be said to be significant are highlighted.}
\centering
\begin{tabular}{@{}ccc|ccc@{}}
\toprule
\textbf{Family} & \textbf{Model} & \textbf{Init.}
& H@3 & $\Omega_{\text{base}}$ & $\Omega_{\text{new}}$ \\
\midrule
\multirow{6}{*}{Transl.}
& TransH & Rand. & 0.272 $\pm$ 0.019 & 0.809 $\pm$ 0.059 & \cellcolor{lightgray} 0.471 $\pm$ 0.067 \\
& TransH & Sch.  & \textbf{0.322 $\pm$ 0.010} & \textbf{0.901 $\pm$ 0.026} & \textbf{0.516 $\pm$ 0.028} \\
\cmidrule{2-6}
& RotatE & Rand. & 0.212 $\pm$ 0.020 & 0.567 $\pm$ 0.028 & 0.231 $\pm$ 0.031 \\
& RotatE & Sch.  & \textbf{0.351 $\pm$ 0.008} & \textbf{0.894 $\pm$ 0.014} & \textbf{0.279 $\pm$ 0.039} \\
\cmidrule{2-6}
& TransR & Rand. & 0.152 $\pm$ 0.011 & 0.719 $\pm$ 0.028 & \textbf{0.254 $\pm$ 0.015} \\
& TransR & Sch.  & \textbf{0.213 $\pm$ 0.005} & \textbf{0.901 $\pm$ 0.026} & 0.245 $\pm$ 0.026 \\
\midrule
\multirow{6}{*}{Semant.}
& DistMult & Rand. & 0.268 $\pm$ 0.007 & 0.782 $\pm$ 0.016 & \cellcolor{lightgray} 0.257 $\pm$ 0.030 \\
& DistMult & Sch.  & \textbf{0.330 $\pm$ 0.005} & \textbf{0.942 $\pm$ 0.011} & \textbf{0.274 $\pm$ 0.013} \\
\cmidrule{2-6}
& HolE & Rand.     & \cellcolor{lightgray} 0.252 $\pm$ 0.014 & \cellcolor{lightgray} 0.899 $\pm$ 0.027 & 0.127 $\pm$ 0.023 \\
& HolE & Sch.      & \textbf{0.254 $\pm$ 0.022} & \textbf{0.915 $\pm$ 0.044} & \textbf{0.317 $\pm$ 0.023} \\
\cmidrule{2-6}
& ProjE & Rand.     & 0.351 $\pm$ 0.000 & \textbf{0.994 $\pm$ 0.001} & 0.176 $\pm$ 0.007 \\
& ProjE & Sch.      & \textbf{0.352 $\pm$ 0.001} & \textbf{0.994 $\pm$ 0.001} & \textbf{0.238 $\pm$ 0.009} \\
\bottomrule
\end{tabular}
\label{tab:kges}
\end{table*}

\begin{table*}[ht]
\caption{Fine-tuning performance on \textbf{FBinc-M} dataset (mean $\pm$ std). In gray, the metrics for which \textit{Schema} initialization has an average greater performance but cannot be said to be significant are highlighted.}
\centering
\begin{tabular}{@{}ccc|ccc@{}}
\toprule
\textbf{Family} & \textbf{Model} & \textbf{Init.}
& H@3 & $\Omega_{\text{base}}$ & $\Omega_{\text{new}}$ \\
\midrule
\multirow{6}{*}{Transl.}
& TransH & Rand. & 0.314 $\pm$ 0.016 & \cellcolor{lightgray} 0.909 $\pm$ 0.026 & 0.236 $\pm$ 0.026 \\
& TransH & Sch.  & \textbf{0.343 $\pm$ 0.009} & \textbf{0.944 $\pm$ 0.022} & \textbf{0.344 $\pm$ 0.008} \\
\cmidrule{2-6}
& RotatE & Rand. & 0.266 $\pm$ 0.007 & 0.633 $\pm$ 0.027 & 0.203 $\pm$ 0.013 \\
& RotatE & Sch.  & \textbf{0.282 $\pm$ 0.015} & \textbf{0.796 $\pm$ 0.015} & \textbf{0.246 $\pm$ 0.009} \\
\cmidrule{2-6}
& TransR & Rand. & 0.139 $\pm$ 0.004 & 0.645 $\pm$ 0.038 & \textbf{0.198 $\pm$ 0.025} \\
& TransR & Sch.  & \textbf{0.217 $\pm$ 0.008} & \textbf{0.936 $\pm$ 0.045} & 0.193 $\pm$ 0.017 \\
\midrule
\multirow{6}{*}{Semant.}
& DistMult & Rand. & 0.317 $\pm$ 0.008 & 0.921 $\pm$ 0.019 & 0.146 $\pm$ 0.030 \\
& DistMult & Sch.  & \textbf{0.346 $\pm$ 0.001} & \textbf{0.973 $\pm$ 0.003} & \textbf{0.232 $\pm$ 0.004} \\
\cmidrule{2-6}
& HolE & Rand.     & 0.263 $\pm$ 0.004 & 0.957 $\pm$ 0.017 & 0.094 $\pm$ 0.020 \\
& HolE & Sch.      & \textbf{0.303 $\pm$ 0.008} & \textbf{0.995 $\pm$ 0.006} & \textbf{0.280 $\pm$ 0.011} \\
\cmidrule{2-6}
& ProjE & Rand.     & 0.318 $\pm$ 0.003 & \cellcolor{lightgray} 0.964 $\pm$ 0.007 & 0.181 $\pm$ 0.005 \\
& ProjE & Sch.      & \textbf{0.326 $\pm$ 0.003} & \textbf{0.966 $\pm$ 0.007} & \textbf{0.247 $\pm$ 0.001} \\
\bottomrule
\end{tabular}
\label{tab:kgem}
\end{table*}

\begin{table*}[ht]
\caption{Fine-tuning performance on \textbf{FBinc-L} dataset (mean $\pm$ std). In gray, the metrics for which \textit{Schema} initialization has an average greater performance but cannot be said to be significant are highlighted.}
\centering
\begin{tabular}{@{}ccc|ccc@{}}
\toprule
\textbf{Family} & \textbf{Model} & \textbf{Init.}
& H@3 & $\Omega_{\text{base}}$ & $\Omega_{\text{new}}$ \\
\midrule
\multirow{6}{*}{Transl.}
& TransH & Rand. & 0.268 $\pm$ 0.014 & 0.809 $\pm$ 0.041 & 0.260 $\pm$ 0.016 \\
& TransH & Sch.  & \textbf{0.333 $\pm$ 0.009} & \textbf{0.918 $\pm$ 0.019} & \textbf{0.328 $\pm$ 0.009} \\
\cmidrule{2-6}
& RotatE & Rand. & 0.300 $\pm$ 0.005 & 0.714 $\pm$ 0.008 & 0.359 $\pm$ 0.009 \\
& RotatE & Sch.  & \textbf{0.310 $\pm$ 0.006} & \textbf{0.735 $\pm$ 0.009} & \textbf{0.431 $\pm$ 0.012} \\
\cmidrule{2-6}
& TransR & Rand. & 0.086 $\pm$ 0.006 & 0.438 $\pm$ 0.032 & \cellcolor{lightgray} 0.232 $\pm$ 0.006 \\
& TransR & Sch.  & \textbf{0.165 $\pm$ 0.005} & \textbf{0.717 $\pm$ 0.023} & \textbf{0.246 $\pm$ 0.015} \\
\midrule
\multirow{6}{*}{Semant.}
& DistMult & Rand. & 0.305 $\pm$ 0.004 & 0.945 $\pm$ 0.008 & 0.148 $\pm$ 0.004 \\
& DistMult & Sch.  & \textbf{0.331 $\pm$ 0.006} & \textbf{0.975 $\pm$ 0.013} & \textbf{0.223 $\pm$ 0.004} \\
\cmidrule{2-6}
& HolE & Rand.     & 0.235 $\pm$ 0.007 & 0.938 $\pm$ 0.028 & 0.110 $\pm$ 0.007 \\
& HolE & Sch.      & \textbf{0.277 $\pm$ 0.010} & \textbf{0.971 $\pm$ 0.018} & \textbf{0.241 $\pm$ 0.008} \\
\cmidrule{2-6}
& ProjE & Rand.     & 0.271 $\pm$ 0.001 & \textbf{0.919 $\pm$ 0.001} & 0.134 $\pm$ 0.001 \\
& ProjE & Sch.      & \textbf{0.281 $\pm$ 0.001} & \textbf{0.919 $\pm$ 0.002} & \textbf{0.190 $\pm$ 0.001} \\
\bottomrule
\end{tabular}
\label{tab:kgel}
\end{table*}